\title{LEARNING COUPLED SPATIAL-TEMPORAL ATTENTION FOR SKELETON-BASED ACTION RECOGNITION}
\documentclass{article}
\usepackage{spconf,amsmath,epsfig,amsmath,cite,hyperref}
\usepackage{graphicx} 
\usepackage{float}
\usepackage{epstopdf}
\usepackage{amsfonts}
\usepackage{url}
\usepackage{flushend}
\begin{document}\sloppy
\def\x{{\mathbf x}}
\def\L{{\cal L}}

\title{LEARNING COUPLED SPATIAL-TEMPORAL ATTENTION FOR SKELETON-BASED ACTION RECOGNITION}
%
\name{Jiayun Wang$^{1}$}\address{\normalsize$^1$School of Electronic and Information Engineering, Beihang University, Beijing,  100191, China}


\maketitle
\begin{abstract}
In this paper, we propose a coupled spatial-temporal attention (CSTA) model for skeleton-based action recognition,
which aims to figure out the most discriminative joints and frames in spatial and temporal domains simultaneously.
Conventional approaches usually consider all the joints or frames in a skeletal sequence equally important, 
which are unrobust to ambiguous and redundant information.
To address this,
we first learn two sets of weights for different joints and frames through two subnetworks respectively,
which enable the model to have the ability of ``paying attention to'' the relatively informative section.
Then, we calculate the cross product based on the weights of joints and frames for the coupled spatial-temporal attention.
Moreover, our CSTA mechanisms can be easily plugged into existing hierarchical CNN models (CSTA-CNN) to realize their function. 
Extensive experimental results on the recently collected UESTC dataset and the currently largest NTU dataset have shown the effectiveness of our proposed method for skeleton-based action recognition.

\end{abstract}

\newcommand{\RNum}[1]{\uppercase\expandafter{\romannumeral #1\relax}}

\section{Introduction}
\label{sec:intro}
Human action recognition is one of the most practical and important branches in the field of computer vision. With the rapid development of this field, various applications such as human-computer interaction, virtual reality, intelligent video surveillance and intelligent house system have been facilitated. 3D skeleton data, compared with traditional RGB videos, has more abstract and compact information, which is more robust to background noise~\cite{DBLP:journals/cviu/HanRHZ17}. Besides, due to its small space usage, lightweight model and short-time training become possible. Therefore, more and more researches have focused on 3D skeleton-based sequences~\cite{DBLP:conf/cvpr/XiaCA12,DBLP:conf/ijcai/GowayyedTHE13,
DBLP:conf/eccv/WangYHLZ16,JunwuCVPR17,Du_2015_CVPR,Mahasseni_2016_CVPR,Zhu2016Co,Song2016An,Liu2016Spatio,Ke2017A,Liu2017Enhanced,Wang2017Modeling,DBLP:journals/corr/ZhangLXZXZ17,Li2017Skeleton,Wang2017Skeleton,tysdeep,Yang2018Action,Li2018Co}.

\begin{figure}[t]
\centering
\includegraphics[width=0.91\linewidth]{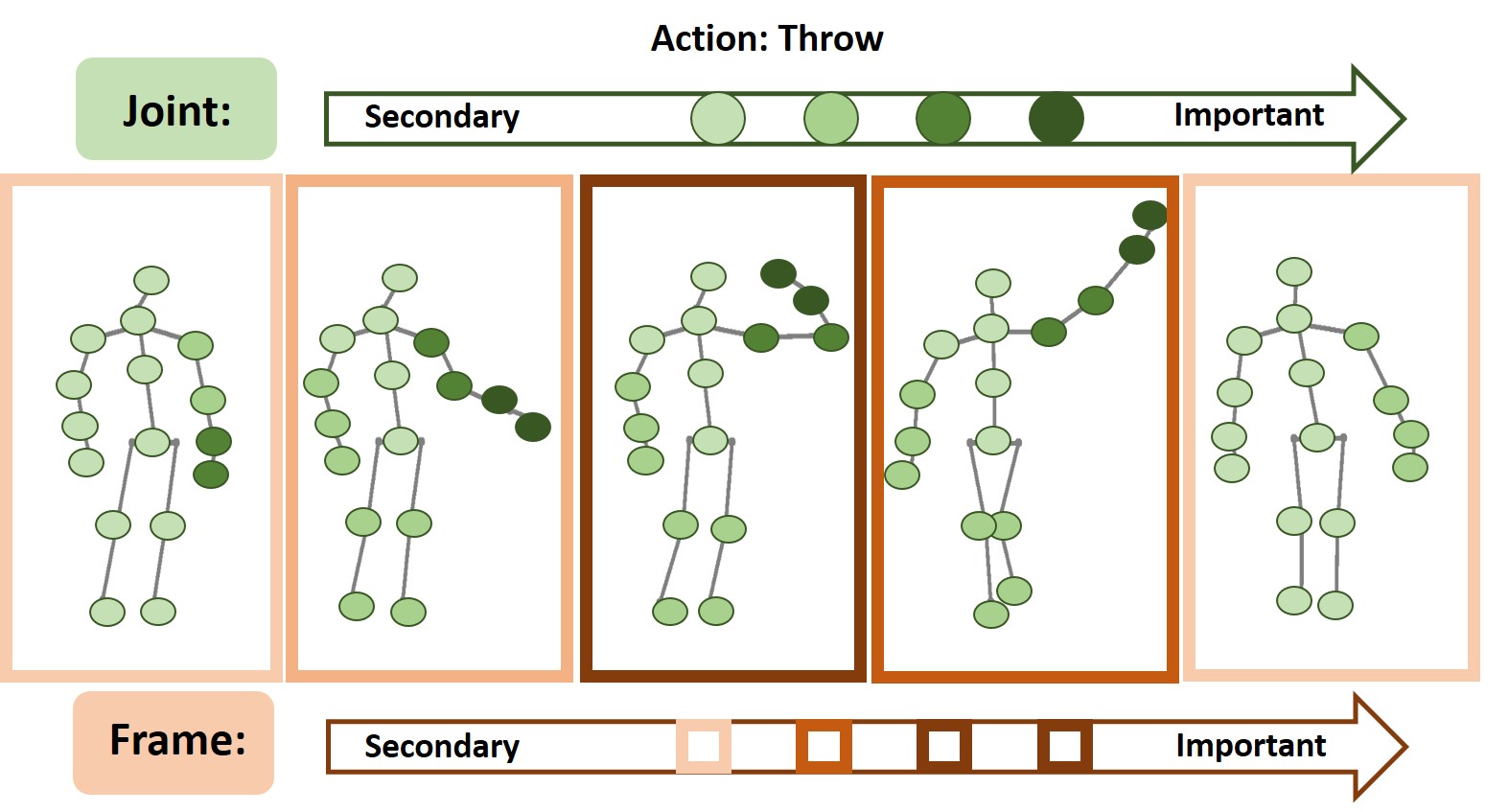}  
\caption{\textbf{The schematic diagram. } 
For the action ``throw'', the frame describing throwing out and the joints of left hands and arms are more informative than others.
To distill these discriminative frames in temporal domain and joints in spatial domain is the key for action recognition task. }
\label{fig:picture001}  
\end{figure}

\begin{figure*}[t]
\centering
\includegraphics[width=0.91\linewidth]{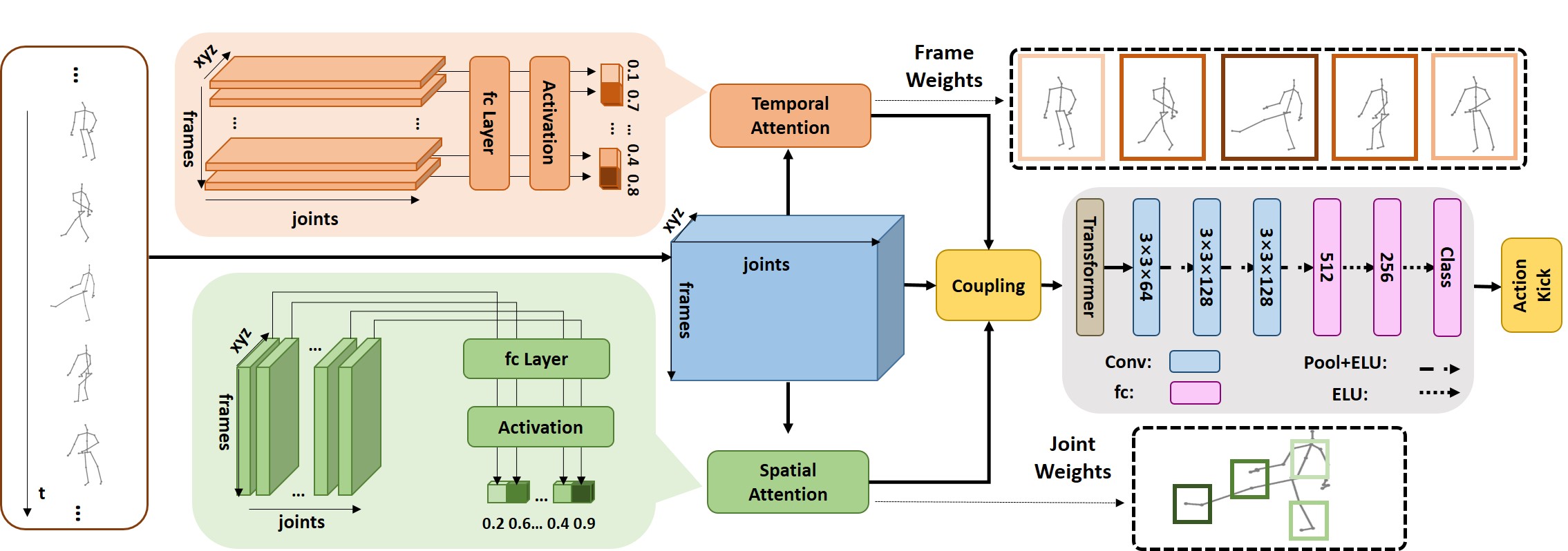}
\caption{\textbf{An overview of our CSTA-CNN model. } The input of our model is a skeleton sequence with the size of $T \times N \times 3$.
The original input is first fed into the spatial and temporal attention subnetworks, which produce a series of frame weights and joint weights. Then, the input is interacted with two set of weights by means of coupling: the cross product of two set of weights operates with the inputs by element-wise multiplication. Finally, the data flow passes through several convolutional layers for feature extraction and fully connected layers for final classification.}
\label{fig:picture002}  
\end{figure*}

There have been various approaches for skeleton-based action recognition, which can be divided into the hand-crafted based~\cite{DBLP:conf/cvpr/XiaCA12,DBLP:conf/ijcai/GowayyedTHE13,
DBLP:conf/eccv/WangYHLZ16,JunwuCVPR17} and deep learning based~\cite{Du_2015_CVPR,Mahasseni_2016_CVPR,Zhu2016Co,Song2016An,Liu2016Spatio,Ke2017A,Liu2017Enhanced,Wang2017Modeling,DBLP:journals/corr/ZhangLXZXZ17,Li2017Skeleton,Wang2017Skeleton,tysdeep,Yang2018Action,Li2018Co}.
With the growth of convenient and cost-effective human skeleton capture sensors (e.g. Kinect), a number of large-scale 3D skeleton datasets have been created during recent years~\cite{DBLP:conf/cvpr/Firman16,DBLP:journals/pr/ZhangLOWT16}, which have promoted the development of deep learning based methods~\cite{Du_2015_CVPR,Mahasseni_2016_CVPR,Zhu2016Co,Song2016An,Liu2016Spatio,Ke2017A,Liu2017Enhanced,Wang2017Modeling,DBLP:journals/corr/ZhangLXZXZ17,Li2017Skeleton,Wang2017Skeleton,tysdeep,Yang2018Action,Li2018Co} for skeleton-based action recognition.
For example, deep models using recurrent neural network (RNN)~\cite{Du_2015_CVPR,Mahasseni_2016_CVPR,Zhu2016Co,Song2016An,Liu2016Spatio,Wang2017Modeling,DBLP:journals/corr/ZhangLXZXZ17}, especially long-short term memory (LSTM), are powerful of extracting temporal features from skeleton data. Convolutional neural network (CNN) models~\cite{Ke2017A,Liu2017Enhanced,Li2017Skeleton,Yang2018Action,Li2018Co}, on the other hand, could capture both the local spatial features and long-term temporal features, which achieve more promising performance currently. 

For a skeleton image data sample, it records the whole information in spatial and temporal domains. However, the most distinct feature to distinguish between one movement and another movement usually exists in a limited time scale and several key joints. Take the action ``throw'' in Figure 1 as an example, the most important frame is when the person exactly throwing out the thing. Similarly, the most important joints are the arm and hand joints. However, for most skeleton-based action recognition models, all frames and joints are consider to have the same contribution to a recognition task, 
which are unstable to indeterminate and redundant skeleton data.
Though some works have address this issue, the attention information is explore only on the temporal domain~\cite{tysdeep}, 
or spatial domain and temporal domain successively~\cite{Song2016An}, which have limitation to embody the coupled attention in spatial and temporal domains well.

To address this, we propose a hierarchical model with coupled spatial-temporal attention mechanisms as shown in Figure 2. We design a spatial attention subnetwork and a temporal attention subnetwork respectively, which are used to learn the weights for each joint and frame. Through cross product operation, we obtain the coupling of spatial and temporal attention. Besides, in order to take full advantage of skeleton data, we adopt two data augmentation methods to enhance the performance of action recognition. Both quantitative and qualitative experimental results on two large-scale datasets have shown the effectiveness of our proposed method for skeleton-based action recognition.

\section{Approach}
\subsection{Model Overview}
Figure 2 shows the framework of our CSTA-CNN model, which contains two parts: the CSTA network and the baseline model. The spatial and temporal attention subnetworks receive the same original inputs respectively, generating a set of joint weights and frame weights. By calculating the cross product of them, we obtain the spatial-temporal feature map of skeleton data. Through element-wise multiplication, we combine the attention mechanisms into the original input. Finally, the data would pass through the baseline model used for skeleton-based action recognition, which contains a CNN subnetwork for feature extraction and several fully connected layers for the final classification.

\subsection{Coupled Spatial-temporal Attention Network}
We design an CSTA network to enable the model to concentrate more on key frames and joints. The implementation of CSTA network is as follows. We adopt a compact tensor $X \in R^{T \times N \times 3}$ to describe a skeleton-based sequence,
where $T$ is the amount of frames, $N$ is the amount of major body joints and 3 is the $xyz$ 3D dimensional coordinates. 

For the spatial attention subnetwork, we first reshape $X$ into another representation $Y = [y_1, y_2, ..., y_N]$, where $y_i \in R^{3T}$. Then, we employ a fully-connected layer to generate corresponding spatial weights $s_i$ for each joint as:
\begin{eqnarray}
s_i = \sigma (W^s \times y_i + b_i^s), \quad i = 1, 2, ... , N,
\end{eqnarray}
where $\sigma(.)$ denotes the activation function, $W^s$ and $b_i^s$ are the corresponding weighted matrix and bias term for spatial attention learning. The total spatial attention
is denoted as $s_{att} = [s_1,s_2,...s_N]$.

Similarly, for the temporal attention network, the input $X$ is transformed into $Z = [z_1, z_2, ..., z_T]$, where $z_i \in R^{3N}$ The total temporal attention $t_{att} = [t_1,t_2,...t_T]$ is generated through the fully-connected layer and activation function. 

After that, we calculate the cross product of the total spatial and temporal attentions to get the coupled spatial-temporal attention map $M \in R^{T \times N}$:
\begin{eqnarray}
M = t_{att}^T \otimes s_{att}
\end{eqnarray}

In order to match the shape of the input $X \in R^{T \times N \times 3}$, we copy the attention map three times on $xyz$ coordinates to get the final attention map $N \in R^{T \times N \times 3}$. Finally, we attach the attention on the input data and obtain the output $O \in R^{T \times N \times 3}$ through element-wise multiplication:
\begin{eqnarray}
O = N \circ X
\end{eqnarray}

\subsection{Hierarchical CNN Model}
The output $O$ of CSTA network is then fed into a hierarchical CNN model, which contains a skeleton transformer~\cite{Li2017Skeleton} and three convolutional layers, to get the final output $U$. Skeleton transformer could converts original $N$ joints into interpolated $M$ joints through an linear transformation, aiming at depicting the linkage between joints. 

Besides, we adopt two-stream CNN methods~\cite{Li2017Skeleton} to improve performance. The inputs of two-stream network include two types: position $X \in R^{T \times N \times 3}$ and motion $X^{'} \in R^{T \times N \times 3}$. The motion data is the difference between the two adjacent position data in temporal domain. Given the position data $S=\{J_1,J_2...,J_N\}$, the motion data could be represented as $M=S^{t+1}-S^{t}=\{J^{t+1}_{1}-J^{t}_{1},J^{t+1}_{2}-J^{t}_{2},...,J^{t+1}_{N}-J^{t}_{N}\}$, where $J=(x,y,z)$ is the $xyz$ dimensional coordinates of the joints. As shown in Figure 2, the above-mentioned CSTA network and hierarchical CNN model are described with the premise that the input is the position data $X \in R^{T \times N \times 3}$. Similarly, the motion data $X^{'}$ $\in R^{T \times N \times 3}$ could also pass through the CSTA network and hierarchical CNN model to generate the final output $U^{'}$. Finally, $U$ and $U^{'}$ are concatenated and fed into three fc layers for the final classification.

\subsection{Data Augmentation}
In order to improve the diversity of the data, we adopt temporal random sampling~\cite{TSN2016ECCV}, which is first used for RGB images. Besides, we adopt temporal random cropping ~\cite{Li2018Co} to focus on different local data distributions. Concretely, we randomly crop the skeleton sequences and uniformly sample the cropped sequences.

\section{Experiment}
\subsection{Datasets and Implementation Details}
\textbf{UESTC RGB-D Dataset (UESTC)~\cite{'ji2018mm'}:}
The UESTC RGB-D dataset is a recently collected large-scale dataset for action recognition. The dataset contains 25600 samples performed by 118 subjects and captured by 8 fixed views and 1 varying view. All samples are divided into 40 classes. We choose cross-subject (CS) and cross-view \RNum{2} (CV\RNum{2}) in~\cite{'ji2018mm'} as criteria.

\noindent\textbf{NTU RGB+D Dataset (NTU)~\cite{Shahroudy2016NTU}:}
The NTU RGB+D dataset is currently the largest dataset for 3D action recognition, which contains 56880 samples divided into 60 classes captured by 3 cameras and performed by 40 subjects. For evaluation, we follow the cross-subject (CS) and cross-view (CV) settings in~\cite{Shahroudy2016NTU}.

\noindent\textbf{Implementation Details:}
In our model, the inputs have 25 joints, 30 frames and 3 dimensional coordinates $xyz$. For random temporal sampling, we randomly select 30 frames from all in the raw data. For random temporal cropping, we randomly crop a sub-series with the ratio between [0.5, 1] on UESTC dataset and [0.9, 1] on NTU dataset from the whole time series. After that, we uniformly select 30 frames from the cropped series. In our experiments, we do the data augmentation 8 times (4 times for random temporal sampling and 4 times for random temporal cropping) for each sample.

\begin{table}[t]
\caption{\small Comparing of the action recognition accuracy (\%) on the UESTC dataset. Partial results are reported from~\cite{'ji2018mm'}. CS: cross-subject, CV\RNum{2}: cross-view \RNum{2}.} \label{tab:table1}
\vskip 0.1 in
\centering 
\newcommand{\tabincell}[2]{\begin{tabular}{@{}#1@{}}#2\end{tabular}}
\begin{tabular}{|c|c|c|c|}
\hline
Method & CS & CV\RNum{2} & Year  \\
\hline \hline
P-LSTM\cite{Shahroudy2016NTU} &60.0\% & 33.0\% & 2016\\ \hline
TCN\cite{Lea2017Temporal} &56.0\% & 43.0\% & 2017\\ \hline
ST-GCN\cite{Yan2018Spatial} &71.0\% & 56.0\% & 2018\\ \hline
VS-CNN\cite{'ji2018mm'} &76.0\% & 71.0\% & 2018\\
\hline\hline
CSTA-CNN(Ours) & \textbf{95.6\%}&  \textbf{81.1\%} & 2019\\
\hline
\end{tabular}
\end{table}

\begin{table}[t]
\centering 
\newcommand{\tabincell}[2]{\begin{tabular}{@{}#1@{}}#2\end{tabular}}
\caption{\small Comparing of the action recognition accuracy (\%) on the NTU dataset. CS: cross-subject, CV: cross-view.} \label{tab:table2}
\vskip 0.1 in
\begin{tabular}{|c|c|c|c|}
\hline
Method & CS & CV & Year  \\
\hline \hline
Part-aware LSTM \cite{Shahroudy2016NTU} & 62.9\%   &   70.3\% & 2016\\
\hline
ST-LSTM+Trust Gate \cite{Liu2017Skeleton} & 69.2\%   &   77.7\% & 2016\\
\hline
LieNet-3Blocks \cite{Huang2016Deep} & 61.4\% & 67.0\% & 2017\\
\hline
Two-Stream RNN \cite{Wang2017Modeling} & 71.3\% & 79.5\% & 2017\\
\hline
STA-LSTM \cite{Song2016An} & 73.4\%   &   81.2\% & 2017\\
\hline
VA-LSTM \cite{DBLP:journals/corr/ZhangLXZXZ17} & 79.2\%   &   87.7\% & 2017\\
\hline
Two-Stream CNN\cite{Li2017Skeleton} \ & 83.2\% & 89.3\% & 2017\\
\hline
LSTM-CNN\cite{Wang2017Skeleton} & 82.9\% & 90.1\% & 2017\\
\hline
TSSI + GLAN + SSAN\cite{Yang2018Action} & 82.4\% & 89.1\%  & 2018\\
\hline
DPRL + GCNN\cite{tysdeep} & 83.5\% & 89.8\% & 2018\\
\hline 
HCN\cite{Li2018Co} & \textbf{86.5\%} & \textbf{91.1\%} & 2018\\
\hline\hline
CSTA-CNN(Ours)&84.9\% & 89.9\% & 2019 \\ \hline 
\end{tabular}
\end{table}

\begin{figure*}[t]
\centering
\includegraphics[width=0.91\linewidth]{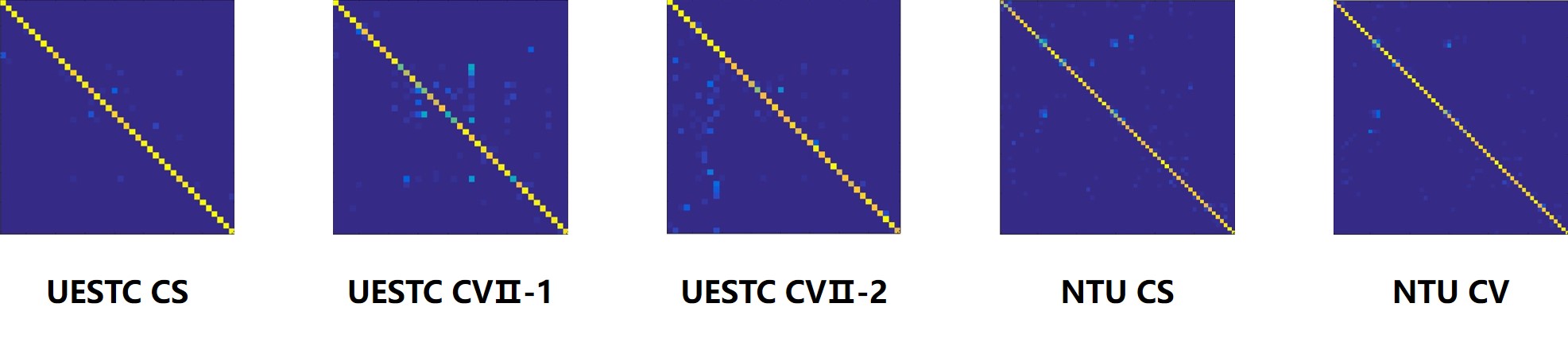}  
\caption{\textbf{The confusion matrix of CSTA-CNN on different settings in UESTC and NTU datasets.}}
\label{fig:picture003}  
\end{figure*}

\subsection{Comparison with State-of-the-arts}
\textbf{UESTC RGB-D Dataset:}
On the newly published large-scale dataset UESTC, our method achieves the highest accuracy for action classification on both CS and CV\RNum{2} settings according to Table 1. Compared with the state-of-the-art method VS-CNN~\cite{'ji2018mm'}, our method improves 20\% on CS setting and 10\% on CV\RNum{2} setting. Our approach achieves higher performance on CS setting probably because the change between subjects is relatively less than CV setting, leading to the better accuracy.

\noindent\textbf{NTU RGB+D Dataset:} 
The performances of different methods on the NTU dataset are listed in Table 2. CSTA-CNN performs slightly lower compared with HCN~\cite{Li2018Co}, probably because it contains co-occurrence mechanisms, which are effective in extracting global spatial features. However, co-occurrence mechanisms could be complementary with our approach and probably achieve even higher performance. Besides, our approach has explicit physical explanation and could be visualize clearly as shown in Figure 4. Specifically, compared with Two-Stream CNN~\cite{Li2017Skeleton}, the idea of which is adopted in our hierarchical CNN model, our approach has 1.7\% and 0.6\% accuracy gain on CS and CV setting, demonstrating the effectiveness of our attention mechanisms to a certain extent.

\newcommand{\tabincell}[2]{\begin{tabular}{@{}#1@{}}#2\end{tabular}}  
\begin{table}[t]
\caption{\small Ablation experiment on attention mechanisms. S: Spatial attention, T: Temporal attention.} \label{tab:table3}
\vskip 0.1 in
\centering 
\begin{tabular}{|c|c|c|c|c|}
\hline
\tabincell{c}{Attention\\setting} & \tabincell{c}{UESTC\\ CS} & \tabincell{c}{UESTC\\CV\RNum{2}} & \tabincell{c}{NTU\\CS} & \tabincell{c}{NTU\\CV}\\
\hline \hline
CSTA-CNN & \textbf{95.6\%} & \textbf{81.1\%} & \textbf{84.9\%} & \textbf{89.9\%} \\ \hline
Without ST & 95.1\% & 78.4\% & 84.4\% & 88.9\% \\ \hline
Without S &  95.4\% & 79.9\% & 84.3\% & 88.8\% \\ \hline
Without T &  95.2\% & 80.1\% & 84.2\% & 89.1\% \\ \hline
\end{tabular}
\end{table}

\subsection{Analysis of the Coupled Spatial-Temporal Attention Mechanisms}
\textbf{Ablation Study:}
In order to research the function of CSTA mechanisms in our model, an ablation experiment is performed as shown in Table 3. The performances of the model without ST decrease 0.5\%, 1.0\%, 0.5\% and 2.7\% on four different settings, which demonstrates the effectiveness of CSTA mechanisms. For some settings, the accuracy of without S or without T is slightly lower than the accuracy of without ST. That is probably because the model without S simply regards the joint weights are invariant for all frames and the model without T simply regards the frame weights are invariant for all joints, which might affect the final performance. Besides, the improvement on CV settings is more evident than that on CS settings. It is probably because the difference between views is greater than that between subjects, causing the discriminative frames and joints more important in CV settings.

\noindent \textbf{Visualization Results:}
For the sake of exhibit the effect of our coupled attention mechanisms directly, we select two representative actions for visualization in Figure 4. As for the action "wear on glasses", the important joints are mainly located at the two arms and hands of the person, because arms and hands have evident movement and the rest of the body basically remain static when wearing on glasses. Furthermore, the middle frame, when the person is exactly wearing on the glasses, is more important than other frames describing the preparation and the end. The situation of action "falling" is similar to the former and would not be detailed.

\begin{figure}[t]
\centering
\includegraphics[width=0.91\linewidth]{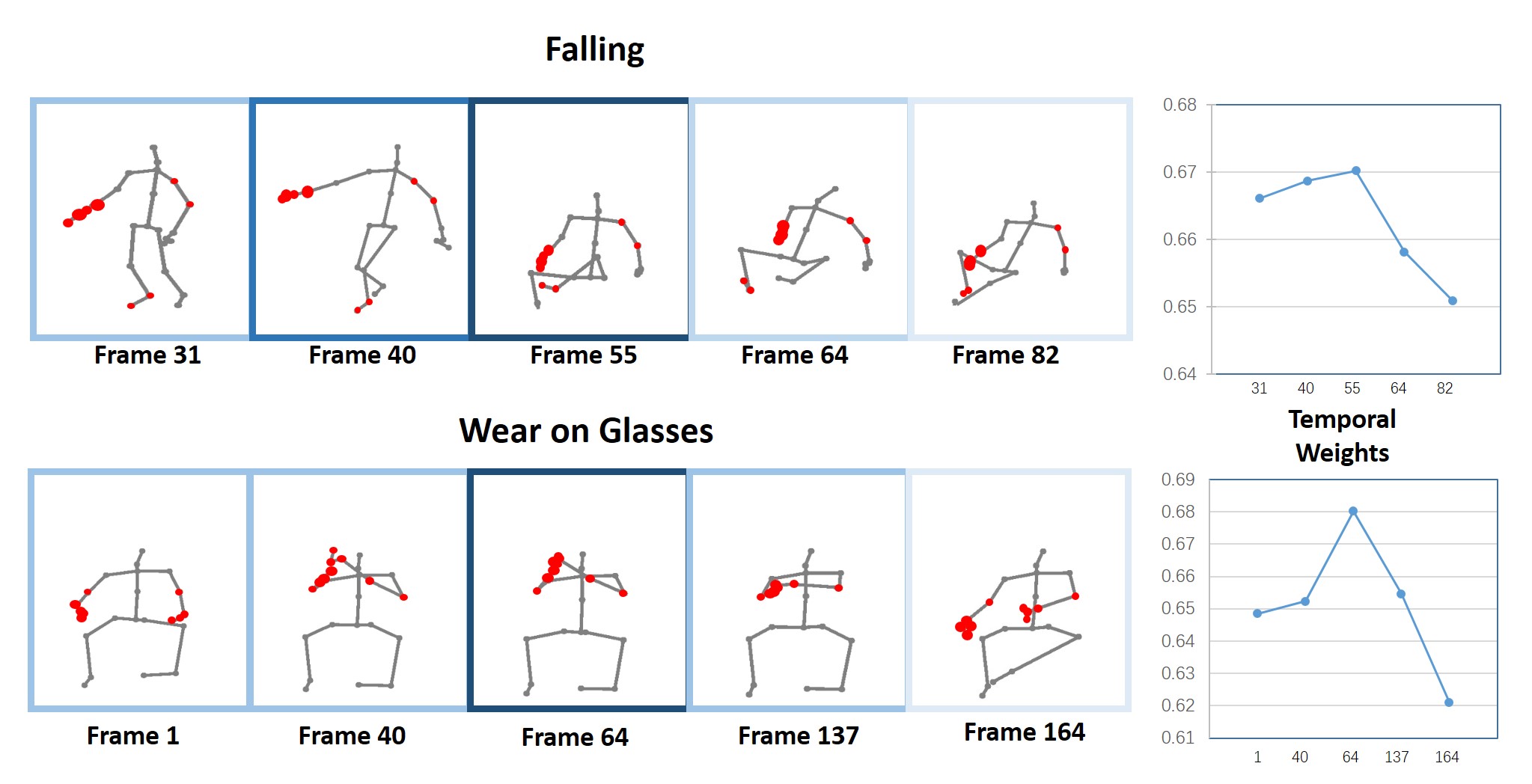}  
\caption{\textbf{The visualization results of coupled spatial-temporal attention mechanisms. }\small The joints with high absolute weight are marked by red and the size is positive correlated with their importance. The importance of the frame weights are exhibited through the borders and the absolute value of them are on the right.}
\label{fig:picture004}  
\end{figure}

\section{Conclusion}
In this paper, we have proposed a multi-layer CNN model with CSTA mechanisms. For the target of concentrating on the most distinct feature on spatial-temporal domain, we have designed two attention subnetworks respectively and combined the outcomes through cross product and element-wise multiplication. Our approach has been systematically evaluated on UESTC and NTU datasets and has achieved promising performance compared with state-of-the-arts. We have also proved the effectiveness of CSTA mechanisms through ablation experiment.

\section{Acknowledgement}
The author would like to sincerely thank Prof. Jiwen Lu and Yansong Tang for their generous support and help to the project. 

\bibliographystyle{IEEEbib}
\footnotesize
\bibliography{icip}
\end{document}